\documentclass[conference]{IEEEtran}
\IEEEoverridecommandlockouts
\usepackage{cite}
\usepackage{amsmath,amssymb,amsfonts}
\usepackage{algorithmic}
\usepackage{graphicx}
\usepackage{subfig}
\usepackage{textcomp}
\usepackage{xcolor}
\usepackage{booktabs}
\usepackage{tabularx}
\usepackage{enumitem}
\usepackage{float}
\usepackage{balance}
\usepackage{hyperref}
\usepackage{pgf-pie}
\usepackage{url}

\usepackage{tikz}
\usetikzlibrary{shapes.geometric, arrows.meta, positioning, fit, backgrounds, calc}
\usepackage{pgfplots}
\pgfplotsset{compat=1.18}
\usepgfplotslibrary{groupplots}

\definecolor{cNaive}{HTML}{D93025}
\definecolor{cStepFn}{HTML}{F9AB00}
\definecolor{cOurs}{HTML}{34A853}
\definecolor{cAlwaysOn}{HTML}{D93025}
\definecolor{cEventDriven}{HTML}{34A853}

\def\BibTeX{{\rm B\kern-.05em{\sc i\kern-.025em b}\kern-.08em
    T\kern-.1667em\lower.7ex\hbox{E}\kern-.125emX}}

\begin{document}

\title{Event-Driven ML Pipeline Orchestration for Manufacturing: An AWS Industry Experience\\
\thanks{\normalfont{Zhengyang ``Cissy'' Gu is the corresponding author (*).}}
\thanks{\normalfont{\S Z. Gu, ORCID: 0009-0007-8564-9907}}
\thanks{\normalfont{\ddag J. E. Hernandez, ORCID: 0000-0002-2327-4034}}
\thanks{\normalfont{This project was funded by Liveline Technologies. Liveline Technologies is dedicated to improving manufacturing performance by harnessing the power of Artificial Intelligence to automate complex processes and predict future problems.}}
\thanks{\normalfont{This is the author's version of this work. It has been accepted for publication in the 14th IEEE International Conference on Cloud Engineering (IC2E 2026). The final published version will be made available on IEEE Xplore.}}
\thanks{\normalfont{\copyright~2026 IEEE. Personal use of this material is permitted. Permission from IEEE must be obtained for all other uses, in any current or future media, including reprinting/republishing this material for advertising or promotional purposes, creating new collective works, for resale or redistribution to servers or lists, or reuse of any copyrighted component of this work in other works.}}
}

\author{
\parbox{\textwidth}{\centering
\parbox[t]{0.3\textwidth}{\centering
{\bfseries Zhengyang ``Cissy'' Gu*\textsuperscript{\S}}\\
{\itshape Data Science, Liveline Technologies}\\
Livonia, MI, USA\\
cissy.gu@liveline.tech}
\hfill
\parbox[t]{0.3\textwidth}{\centering
{\bfseries Thomas Cook}\\
{\itshape Engineering, Liveline Technologies}\\
Livonia, MI, USA\\
thomas.cook@liveline.tech}
\hfill
\parbox[t]{0.3\textwidth}{\centering
{\bfseries Fredaljohn Rohrbaugh}\\
{\itshape Engineering, Liveline Technologies}\\
Livonia, MI, USA\\
fj.rohrbaugh@liveline.tech}
\\[1.5em]
\makebox[\textwidth]{\parbox[t]{0.3\textwidth}{\centering
{\bfseries Joseph E. Hernandez\textsuperscript{\ddag}}\\
{\itshape COO, Liveline Technologies}\\
Livonia, MI, USA\\
joseph.hernandez@liveline.tech}
\hspace{3em}
\parbox[t]{0.3\textwidth}{\centering
{\bfseries Chris Couch}\\
{\itshape CEO, Liveline Technologies}\\
Livonia, MI, USA\\
chris.couch@liveline.tech}}
}
}

\maketitle

\begin{abstract}
We present an industry experience report on three years of operating an event-driven cloud infrastructure for continuous machine learning training in automotive manufacturing. Our system orchestrates GPU-accelerated training of product-specialized model pairs, a physics prediction model and a reinforcement-learning control policy, across multiple plants, coordinating long-running GPU workloads triggered by manufacturing events. The architecture combines Amazon ECS with EC2 GPU capacity providers, SQS-based messaging with dead-letter queues, and an admission-controlled Lambda dispatcher that enforces cluster concurrency limits. A Conductor orchestrator on ECS Fargate initiates dependency-aware retraining chains on a weekly schedule. The entire infrastructure is codified in modular Terraform with multi-account separation. From 40{,}000+ production training jobs we report a 72--78\% cost reduction versus always-on GPU infrastructure. A discrete-event simulation confirms that admission control is necessary (naive dispatch loses 65\% of jobs) and that queue-draining matches AWS Step Functions latency while eliminating per-job startup overhead. We provide lessons learned and release the simulator and Terraform module skeletons as open-source artifacts.
\end{abstract}



\section{Introduction}

Cloud-based ML training infrastructure must balance GPU cost efficiency (instances at \$1--\$10+/hour), long-running job coordination (2--12 hours), fault tolerance, and operational simplicity for small teams~\cite{sculley2015hidden, kreuzberger2023mlops, amershi2019software}. These challenges intensify in manufacturing AI, where \textit{product-specialized models} multiply the training demand: a single plant may require dozens of independently trained models, each triggered by equipment changes, performance degradation, or data drift~\cite{paleyes2022challenges}.

Our key insight is that manufacturing ML training is fundamentally an \textbf{event-driven batch workload}: jobs are triggered by discrete events, require expensive GPU compute for bounded durations, and produce artifacts that flow through evaluation pipelines before deployment. This pattern maps naturally to message queues, serverless dispatchers, and container-based GPU compute, rather than to always-on clusters that idle between bursts or managed platforms whose per-job overhead becomes significant at scale. This paper is an \emph{industry experience report} on building and operating such an infrastructure on AWS for over three years, where the system triggers, executes, evaluates, and promotes models \textit{autonomously} without human intervention for the common case.

\subsection{Problem and Domain}
\label{sec:background}

Manufacturing processes require product-specialized ML models that must be continuously retrained as production conditions change. Our system trains two proprietary model classes per product per plant line: (1)~\textbf{LPP (Liveline Physics Package)}, a sequence-to-sequence network that learns forward dynamics of a manufacturing process from historical sensor data, serving as a digital twin~\cite{tao2019digital}; and (2)~\textbf{LCP (Liveline Controls Package)}, a deep reinforcement-learning control policy trained using the validated LPP as its environment, which recommends setpoints to keep quality signals within specification limits~\cite{spielberg2019toward}.

\begin{figure}[t]
\centering
\includegraphics[width=0.82\columnwidth]{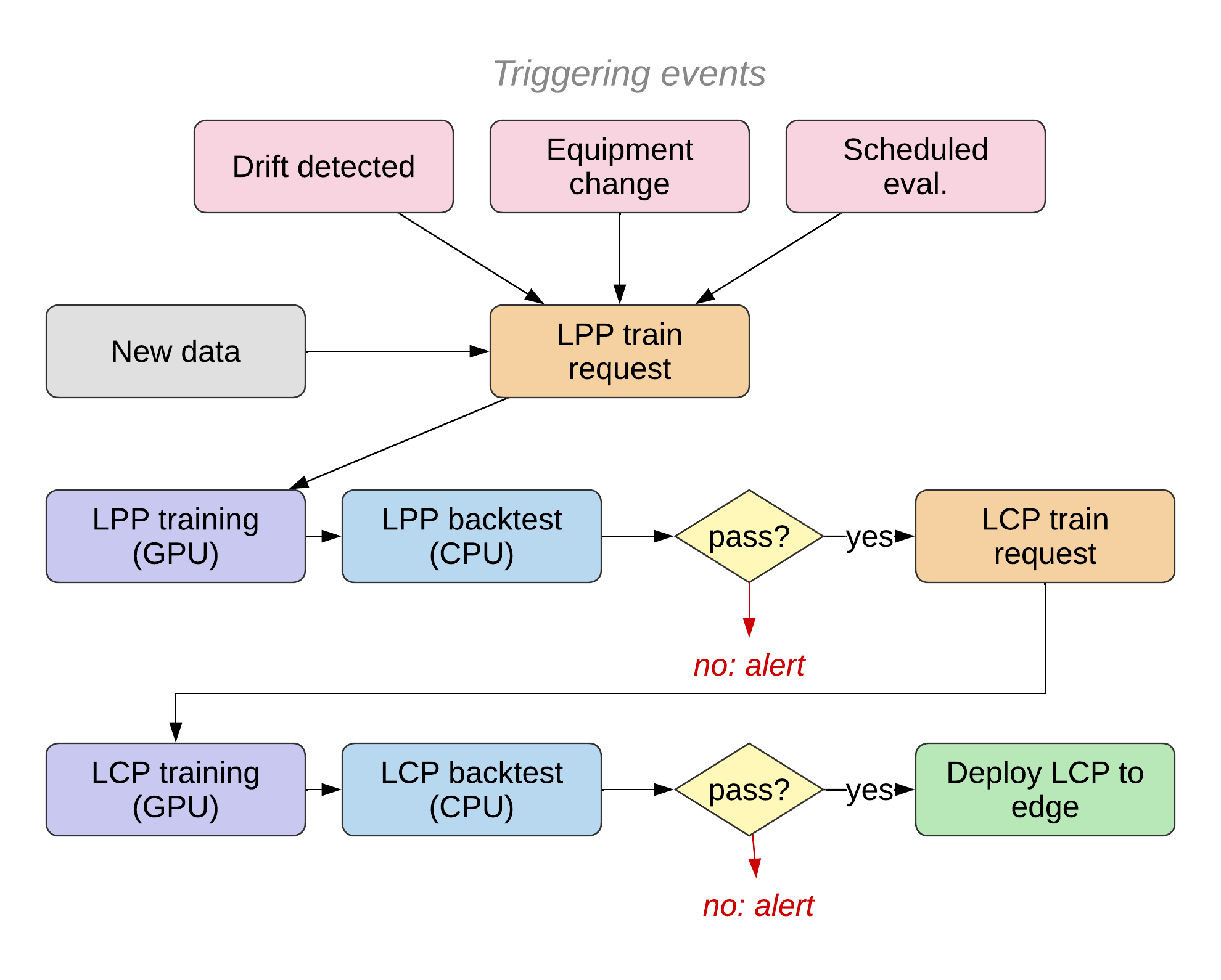}
\caption{End-to-end manufacturing ML workflow per product per plant line. Triggering events initiate an LPP training request with available new data. The LPP physics model trains on GPU and backtests on CPU (Fargate); on pass, the LCP control policy trains on GPU and backtests on CPU; on pass, the LCP deploys to the edge controller. Each transition is decoupled through a durable SQS queue (Section~\ref{sec:dispatcher}).}
\label{fig:workflow}
\end{figure}

Each retraining follows a strict, dependency-aware pipeline (Fig.~\ref{fig:workflow}): a \emph{manufacturing event} (new data, drift, equipment change, scheduled re-evaluation) triggers an LPP training request $\to$ LPP training (GPU, 0.5--1\,h) $\to$ LPP backtest (CPU, 10--20\,min) $\to$ on pass: LCP training (GPU, 1--2\,h) $\to$ LCP backtest (CPU, 10--20\,min) $\to$ on pass: deploy LCP to the edge controller. The LCP cannot be retrained against a stale LPP because its policy validity depends on its twin's fidelity. Each transition is decoupled through a durable SQS queue, allowing failures at any stage to be retried without re-running upstream stages. A \emph{Conductor} orchestrator periodically scans the entire model inventory for each facility, determines which products require retraining, and submits the corresponding requests into this pipeline.

These phases have fundamentally different compute profiles (milliseconds for triggering, hours for training), hardware requirements (CPU-only for orchestration and backtesting, GPU for training), and failure semantics (idempotent retries for training, exactly-once for deployment). This workflow runs concurrently for many products across many plant lines, producing a bursty, GPU-constrained workload pattern.

\subsection{AWS Building Blocks}
\label{sec:aws_background}

Our architecture composes six core AWS services:

\textbf{Amazon SQS} provides durable message queuing with configurable retention (up to 14 days), visibility timeouts, at-least-once delivery, and dead-letter queues (DLQs). SQS buffers burst training request arrivals against bounded GPU capacity, ensuring no request is lost.

\textbf{AWS Lambda} executes code in response to events (including SQS messages) with per-millisecond billing and a 15-minute maximum duration. We use Lambda exclusively for lightweight admission control and dispatch.

\textbf{Amazon ECS} orchestrates Docker containers on two launch types: \emph{Fargate} (serverless) and \emph{EC2} (self-managed, with GPU access). ECS EC2 with capacity providers backed by Auto Scaling Groups enables scale-to-zero GPU clusters.

\textbf{Amazon EventBridge Scheduler} provides cron-based scheduling with timezone support, triggering periodic Conductor scans.

\textbf{AWS Auto Scaling Groups (ASG)} manage EC2 instance fleets integrated with ECS capacity providers, automatically launching GPU instances when tasks are pending and terminating them when idle.

\textbf{Terraform}~\cite{terraform2024} declaratively provisions the entire infrastructure in modular, reusable configurations.

\subsection{Contributions}

\begin{enumerate}[leftmargin=*, nosep]
\item An \textbf{admission-controlled task dispatcher} pattern using Lambda + SQS + ECS that enforces GPU concurrency limits while buffering overflow in durable queues (Section~\ref{sec:dispatcher}).

\item A \textbf{controlled discrete-event simulation} (open-sourced) comparing our pattern against naive dispatch and AWS Step Functions on identical bursty workloads, showing that admission control is necessary (65\% job loss without it) and that queue-draining matches Step Functions latency while eliminating per-job startup overhead (Section~\ref{sec:benchmark}).

\item A \textbf{qualitative architecture comparison} against SageMaker Pipelines, Step Functions, Apache Airflow, and self-managed Kubernetes (Section~\ref{sec:comparison}).

\item \textbf{Three-year operational experience} from 40{,}000+ GPU training jobs including cost analysis (72--78\% reduction), failure modes, and lessons learned (Section~\ref{sec:lessons}).
\end{enumerate}

\textbf{Artifact availability.} The discrete-event simulation, baseline implementations, and a Terraform module skeleton for the dispatcher are released as open-source.\footnote{\url{https://github.com/CissyGu/ic2e-2026-artifacts}}

\section{Related Work}

\subsection{ML Training Infrastructure}

Google's TFX~\cite{baylor2017tfx} and Meta's (formerly Facebook) FBLearner~\cite{hazelwood2018applied} established patterns for production ML platforms at hyperscale. Open-source alternatives, Kubeflow on Kubernetes~\cite{kubeflow2019}, MLflow for experiment tracking~\cite{zaharia2018accelerating} target smaller teams but require significant operational overhead. AWS SageMaker Pipelines~\cite{liberty2020elastic} provides managed orchestration with built-in GPU instance provisioning but imposes per-job startup latency (3--5 minutes for instance warm-up) and pricing premiums. Recent work on domain-aware model training as a service~\cite{zhang2024domain} highlights the need for workload-specific training infrastructure beyond generic platforms.

\subsection{GPU Cluster Scheduling}

Voda~\cite{hsieh2023voda} addresses elastic GPU scheduling in Kubernetes for deep learning, demonstrating that GPU utilization in shared clusters averages only 30--50\%~\cite{burns2016borg}. Optimus~\cite{peng2018optimus} further demonstrates that workload-aware scheduling improves GPU cluster efficiency. Our work complements these by showing that event-driven, queue-buffered GPU provisioning achieves higher effective utilization for batch training workloads by eliminating idle time between jobs entirely. GPU instances exist only when work is queued.

\subsection{Workflow Orchestration Platforms}

Apache Airflow~\cite{airflow2015} is the most widely adopted open-source workflow orchestration platform, using Python-defined DAGs to express task dependencies with rich monitoring and retry logic. However, Airflow's concurrency controls do not integrate with the underlying GPU infrastructure, and its scheduler is always-on regardless of workload. For our bursty pattern (weekly bursts of 50--100 training requests per plant with days of inactivity between), a serverless dispatcher that costs nothing during idle periods is a better economic fit.

\subsection{Event-Driven Cloud Architectures}

Serverless computing and event-driven patterns have been extensively studied for web workloads~\cite{scheuner2020function}. Applications to ML pipelines remain underexplored, partly because training jobs far exceed Lambda's 15-minute limit. Our architecture bridges this gap using Lambda purely for admission control and dispatch, with ECS providing unbounded-duration GPU compute. This hybrid pattern (serverless orchestration with container-based compute) is a growing paradigm in cloud engineering~\cite{barrak2023serverless, jonas2019cloud}.

\subsection{Infrastructure as Code for ML Systems}

Infrastructure-as-code (IaC) tools like Terraform~\cite{terraform2024} enable declarative, version-controlled, and reproducible cloud provisioning, yet Terraform-based ML infrastructure has received limited academic treatment. AWS prescribes \emph{multi-account architectures} separating tooling, compute, and shared-services accounts for least-privilege access~\cite{aws2021secureml}, and documents continuous delivery practices for ML pipelines~\cite{aws2020mlops}. Recent work on CI/CD for Kubernetes~\cite{shrestha2024cicd} and elastic cloud workflows~\cite{dasgupta2024elastic} further demonstrates the community's interest in reproducible infrastructure. Our work contributes a concrete, production-derived, multi-account Terraform reference architecture covering ECS clusters, SQS queues, Lambda dispatchers, and cross-account CI/CD.

\section{System Architecture}

\subsection{Design Requirements}

Our system must satisfy five requirements derived from manufacturing constraints:

\begin{enumerate}[leftmargin=*, nosep, label=\textbf{R\arabic*}]
\item \textbf{Long-running GPU jobs:} Each LPP+LCP training chain takes 1.5--3.5 hours (LPP: 0.5--1\,h GPU + 10--20\,min backtest; LCP: 1--2\,h GPU + 10--20\,min backtest), with 30--50 chains per plant per week, far exceeding all serverless compute limits.
\item \textbf{Event-driven triggering:} Jobs initiated by discrete events (data arrival, schedule, drift), not continuous demand.
\item \textbf{Dependency chains:} LPP training $\to$ LPP backtest $\to$ LCP training $\to$ LCP backtest must execute sequentially per product (Fig.~\ref{fig:workflow}).
\item \textbf{Concurrency control:} GPU instances are expensive; the system must limit concurrent tasks while guaranteeing no job loss.
\item \textbf{Multi-facility autonomy:} Each plant operates independently on its own schedule with facility-specific configurations.
\end{enumerate}

\subsection{Multi-Account AWS Topology}

The infrastructure spans three AWS accounts (Table~\ref{tab:accounts}), following AWS best practices for enterprise ML platforms~\cite{aws2021secureml}. This separation serves two purposes: \emph{least-privilege access} (tooling credentials cannot modify running workloads) and \emph{blast-radius containment} (a misconfigured IAM policy in one account cannot affect resources in another).

\begin{table}[h]
\centering
\caption{Multi-Account AWS Topology}
\label{tab:accounts}
\small
\begin{tabularx}{\columnwidth}{lX}
\toprule
\textbf{Account} & \textbf{Resources} \\
\midrule
Shared Services & Terraform state (S3 + DynamoDB locking). No compute resources. \\
Tooling (CI/CD) & CodePipeline, CodeBuild (build + deploy), ECR base images, VPC, CodeArtifact \\
Customer Compute & 3 ECS clusters, 4 Lambda dispatchers, 8 SQS queues, EventBridge schedules, workload ECR repos, CloudWatch logs \\
\bottomrule
\end{tabularx}
\end{table}

The three accounts communicate through a narrow, controlled interface. When a new code version is ready, CodeBuild in the Tooling account temporarily assumes a \texttt{LambdaDeploymentRole} in the Customer Compute account using \texttt{sts:AssumeRole} with an external ID for verification. This role is restricted to updating Lambda function code and reading from ECR. The deployment flow is: CodeBuild builds a Docker image, pushes it to ECR in the Customer Compute account, and updates the Lambda function to point to the new image. ECS tasks pull the latest image on their next launch. Terraform state for all environments resides in an encrypted S3 bucket in the Shared Services account with DynamoDB-based locking, keeping infrastructure definitions consistent and conflict-free across accounts.

\subsection{Heterogeneous Compute Tiers}
\label{sec:heterogeneous}

Each pipeline stage runs on a different compute tier matched to its resource profile (Table~\ref{tab:compute_tiers}). Rather than using a single orchestrator that runs all stages on the same backend, our system composes three AWS compute services. SQS acts as the integration layer between tiers: each stage writes a completion message to a queue, and the downstream stage consumes from that queue using whichever compute tier is appropriate.

\begin{table}[h]
\centering
\caption{Compute Tier Selection per Pipeline Stage}
\label{tab:compute_tiers}
\small
\begin{tabularx}{\columnwidth}{@{}p{1.0cm}@{\hspace{6pt}}p{1.1cm}@{\hspace{6pt}}p{0.7cm}@{\hspace{6pt}}X@{}}
\toprule
\textbf{Tier} & \textbf{Duration} & \textbf{Cost} & \textbf{Stage and Justification} \\
\midrule
Lambda & $<$1\,s & $<$0.1\% & \textit{Triggering \& dispatch.} SQS-triggered admission control, active-task inspection, ECS:RunTask calls. Pay-per-invocation (\$0 when idle). \\[4pt]
ECS\newline Fargate & 5--30\,min & 1.5\% & \textit{Orchestration (Conductor).} Plant inventory scan, dependency evaluation, SQS message submission. Per-second billing. Scales to zero. \\[4pt]
ECS\newline Fargate & 10--20\,min & 2.5\% & \textit{Backtesting (LPP \& LCP).} CPU-only model inference against held-out data. Per-second billing, no instance management. \\[4pt]
ECS EC2\newline (GPU) & 0.5--2\,h & 96\% & \textit{Training (LPP \& LCP).} \texttt{g4dn.4xlarge} (16\,vCPU, 64\,GB, 1$\times$T4 GPU). Full CUDA/cuDNN stack. ASG capacity provider enables scale-to-zero. \\
\bottomrule
\end{tabularx}
\end{table}

In practice, this means each tier operates at a fundamentally different cost scale. At 30--50 training chains per plant per week (each chain: 1.5--3\,h of GPU time plus 20--40\,min of CPU backtesting):

\begin{itemize}[leftmargin=*, nosep]
\item \textbf{Triggering} (Lambda): $<$1\,s per invocation at \$0.0000002 each. Weekly cost: effectively \$0.
\item \textbf{Orchestration} (Fargate): 5--30\,min per Conductor run at \$0.01--0.05. Runs once per week per plant.
\item \textbf{Backtesting} (Fargate): 10--20\,min per backtest, 60--100 backtests/week (LPP + LCP), at \$0.02--0.05 each. Weekly cost: \$1--5.
\item \textbf{Training} (ECS EC2 GPU): 45--150 GPU-hours/week at \$1.21/h on-demand or \$0.36/h spot. Weekly cost: \$16--180 depending on spot availability.
\end{itemize}

GPU training accounts for 96\% of monthly cost (Table~\ref{tab:compute_tiers}), while all non-GPU stages collectively contribute less than 5\%, validating the heterogeneous-tier approach: each stage pays only for the compute class it actually requires.

\subsection{Admission-Controlled Dispatcher}
\label{sec:dispatcher}

\begin{figure*}[t]
\centering
\includegraphics[width=0.85\textwidth]{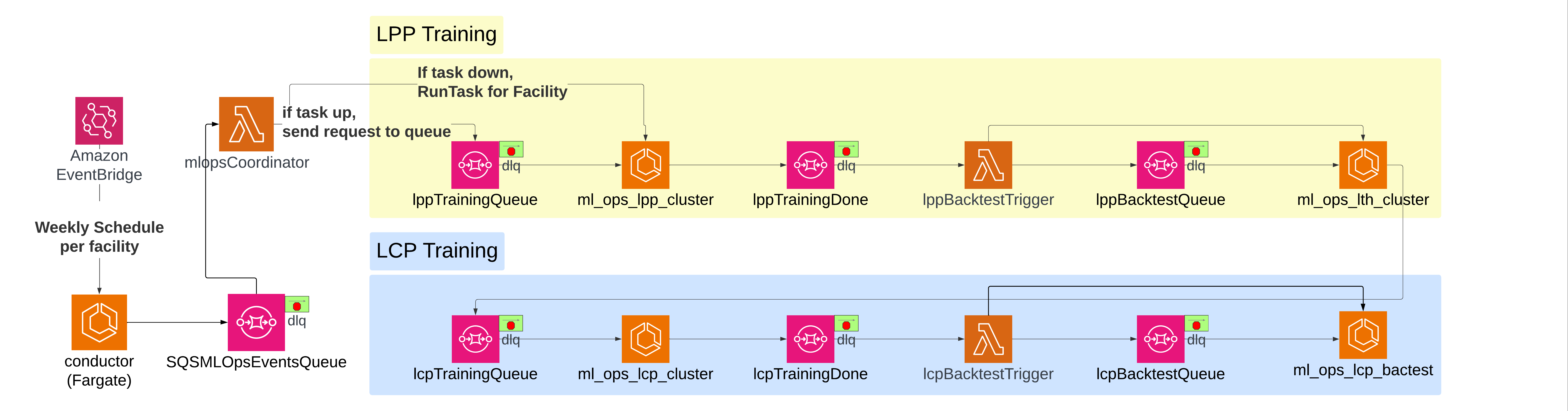}
\caption{AWS-native realization of the workflow in Fig.~\ref{fig:workflow}. EventBridge triggers the Conductor (Fargate), which submits to \texttt{MlopsEventsQueue}. The \texttt{mlopsCoordinator} Lambda enforces concurrency limits before launching GPU tasks. Each stage communicates asynchronously through SQS queues. DLQs capture failures at every stage.}
\label{fig:e2e}
\end{figure*}

The central coordination mechanism is a Lambda-based \textit{admission-controlled task dispatcher} that bridges SQS message queues and ECS GPU/Fargate tasks. This pattern solves the fundamental tension between message-driven triggering (where messages arrive faster than GPU tasks complete) and expensive GPU resource limits. Figure~\ref{fig:e2e} shows how this maps onto the AWS infrastructure: EventBridge Scheduler triggers a Conductor task on ECS Fargate; the Conductor enqueues training requests on \texttt{MlopsEventsQueue}; the admission-controlled Lambda dispatcher gates launching of GPU tasks on ECS EC2; completion events flow forward through additional queues to drive subsequent stages.

\begin{figure}[h]
\centering
\includegraphics[width=0.85\columnwidth]{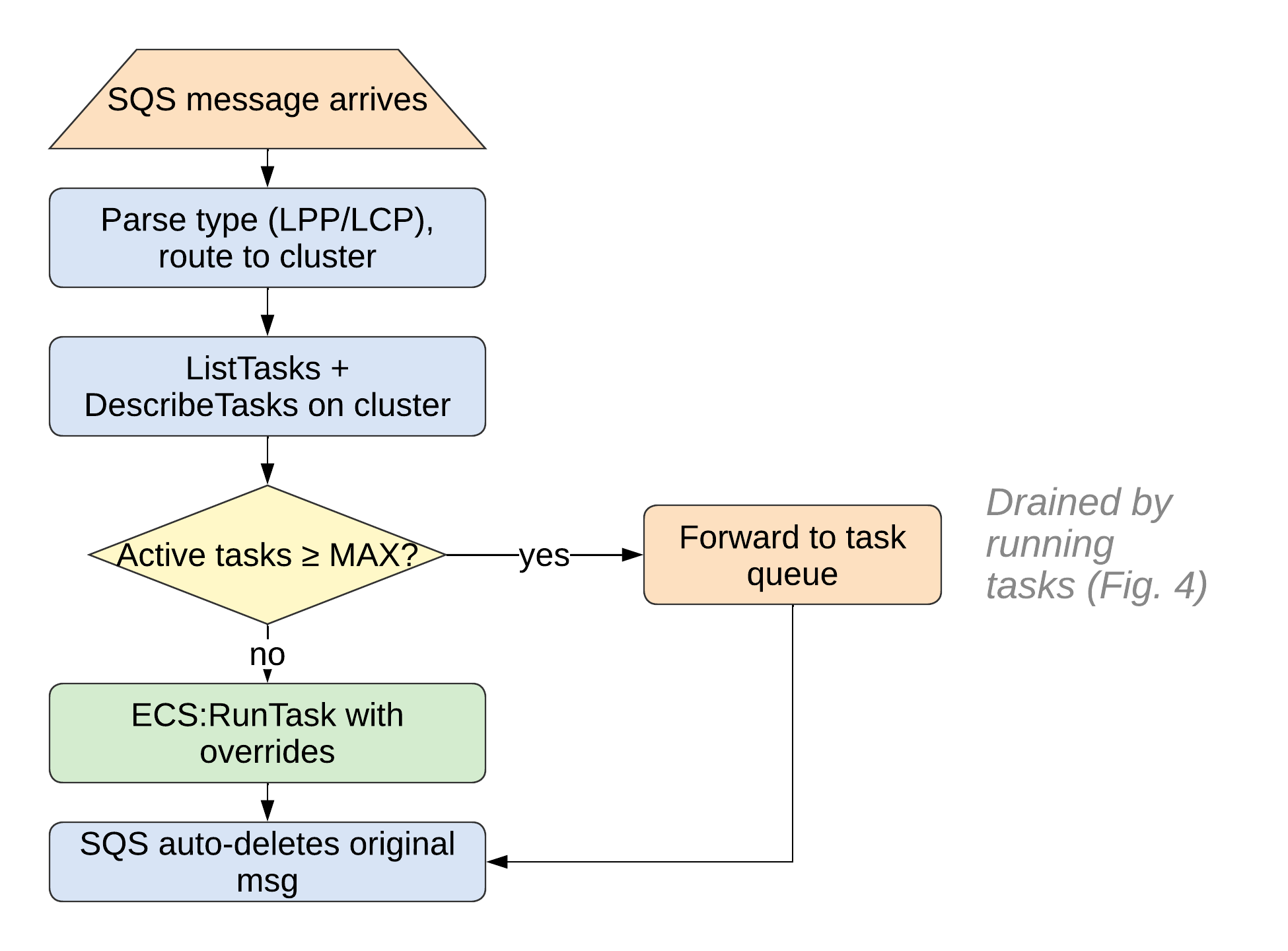}
\caption{Admission-controlled dispatcher logic (runs in Lambda, $<$1\,s). The Lambda parses each SQS message, queries active ECS tasks, and either dispatches a new GPU task or forwards to the task queue for deferred draining by already-running tasks (Fig.~\ref{fig:handler_flow}). In both cases, the original message is auto-deleted from the source queue.}
\label{fig:dispatcher_flow}
\end{figure}

The dispatcher and the handler wrapper (Section~\ref{sec:handler_wrapper}) form two complementary halves of the system: the dispatcher (Fig.~\ref{fig:dispatcher_flow}), running in Lambda for $<$1 second, decides \emph{where} each message goes; the handler wrapper (Fig.~\ref{fig:handler_flow}), running on GPU for hours, determines \emph{how} tasks consume those messages.

The dispatcher Lambda (\texttt{mlopsCoordinator}) is triggered by an SQS event source mapping from \texttt{MlopsEventsQueue}. Each message contains a typed training request specifying the plant, facility ID, request ID, and model type (\texttt{lpp} or \texttt{lcp}). The Lambda routes each message to the appropriate cluster and task definition based on the type field, enabling a single dispatcher to manage multiple heterogeneous training workloads.

For each incoming message, the dispatcher executes the admission control logic shown in Fig.~\ref{fig:dispatcher_flow}:

\textbf{$\bullet$ Parse and route.} Extract the message type (\texttt{lpp}/\texttt{lcp}) to determine the target ECS cluster, task definition family, and task message queue.

\textbf{$\bullet$ Query active tasks.} Call \texttt{ECS:ListTasks} filtered by task family on the target cluster, then \texttt{ECS:DescribeTasks} to count tasks where both \texttt{lastStatus}~$\neq$~\texttt{STOPPED} and \texttt{desiredStatus}~$\neq$~\texttt{STOPPED}.

\textbf{$\bullet$ Check concurrency limit.} If active tasks~$\geq$~\texttt{MAX\_WORKERS} (configured as~2), forward the message to the \emph{task message queue} (e.g., \texttt{lppTrainingQueue}) for deferred processing by running ECS tasks. The original message is still auto-deleted from the source queue because the Lambda returns successfully.

\textbf{$\bullet$ Dispatch if capacity available.} Call \texttt{ECS:RunTask} with container overrides specifying the training handler script and parameters (plant, facility\_id, request\_id).

\textbf{$\bullet$ Automatic cleanup.} In both paths (dispatch or forward), the SQS event source mapping automatically deletes the original message from the source queue upon successful Lambda return.

The dispatcher's concurrency limit (\texttt{MAX\_WORKERS}=2) is deliberately aligned with the ASG's scaling limits (min=0, max=2). Both the Lambda environment variable and the ASG \texttt{max\_size} are derived from the same Terraform variable:

{\small
\begin{verbatim}
variable "lpp_max_workers" {
  description = "Max concurrent LPP GPU tasks"
  type        = number
  default     = 2
}

resource "aws_autoscaling_group" "lpp" {
  max_size = var.lpp_max_workers
  min_size = 0
  ...
}

resource "aws_lambda_function" "coordinator" {
  environment {
    variables = {
      MAX_WORKERS = tostring(var.lpp_max_workers)
    }
  }
}
\end{verbatim}
}

This single-variable alignment is critical: if \texttt{MAX\_WORKERS} exceeds the ASG maximum, GPU instances sit idle; if the ASG maximum exceeds \texttt{MAX\_WORKERS}, dispatched tasks fail with placement errors. By binding both to the same Terraform variable, operators adjust cluster capacity in one place with confidence that the admission controller and infrastructure remain synchronized. The Lambda itself uses \texttt{reserved\_concurrency = 1}, serializing admission decisions and preventing race conditions where concurrent invocations both observe capacity and dispatch simultaneously.

\subsection{Handler Wrapper and Message Draining}
\label{sec:handler_wrapper}

While the dispatcher decides where messages go, the handler wrapper determines how GPU tasks \emph{consume} them. When the dispatcher forwards overflow messages to the task queue (the ``yes'' path in Fig.~\ref{fig:dispatcher_flow}), those messages are picked up by already-running ECS tasks through the draining mechanism described here.

\begin{figure}[h]
\centering
\includegraphics[width=0.82\columnwidth]{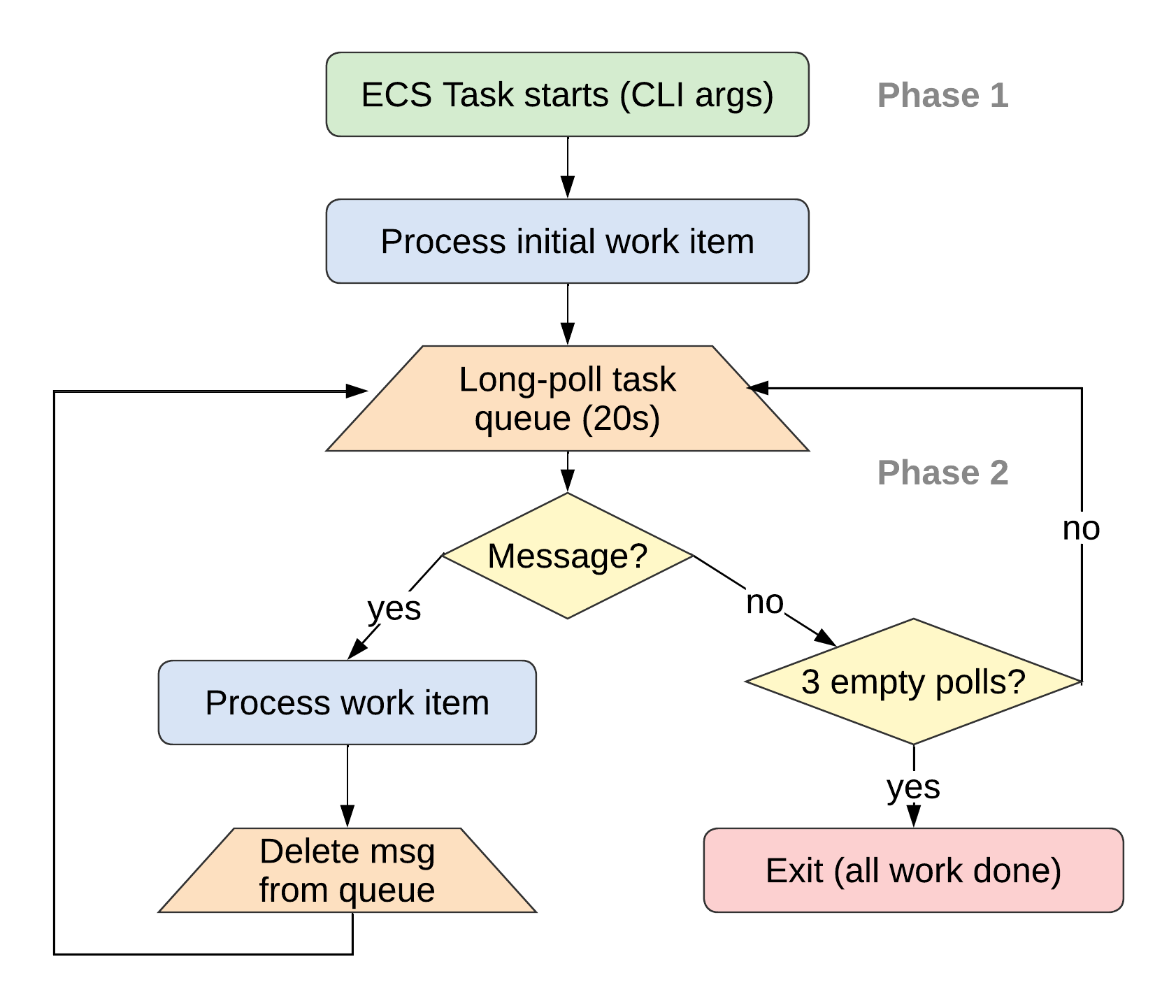}
\caption{ECS task handler wrapper flow. Phase~1 processes the initial work item from CLI arguments. Phase~2 drains the task message queue via long-polling until 3 consecutive empty polls (60s idle), maximizing GPU utilization per launched instance.}
\label{fig:handler_flow}
\end{figure}

The ECS task's entry point is a \textit{handler wrapper} that implements a two-phase execution model (Fig.~\ref{fig:handler_flow}):

\textbf{Phase 1: Initial Work Item.} The handler receives its primary work item via container overrides (CLI arguments: plant, facility\_id, request\_id) injected by the dispatcher Lambda at launch time. It executes this work item immediately without consulting any queue.

\textbf{Phase 2: Queue Draining.} After completing the initial work item, the handler enters a polling loop on the task message queue (\texttt{lppTrainingQueue}), consuming the overflow messages forwarded by the dispatcher. It uses SQS long-polling (\texttt{WaitTimeSeconds=20}) to efficiently wait for messages. For each received message, it processes the training request identically to Phase~1, then deletes the message from the queue. If no messages arrive after 3 consecutive empty polls (60 seconds of silence), the handler concludes that no further work is pending and exits cleanly.

This two-phase pattern enables the system to achieve higher throughput than a simple one-task-per-message model. When bursts of 30--50 requests arrive:
\begin{itemize}[leftmargin=*, nosep]
\item The first 2 messages launch 2 ECS tasks (one per available slot).
\item Subsequent messages are forwarded to the task message queue by the dispatcher.
\item The 2 running ECS tasks drain the overflow messages as they complete their initial work, processing them sequentially without requiring additional ECS task launches.
\end{itemize}

This effectively converts a \texttt{MAX\_WORKERS}-bounded cluster into a throughput-maximizing system: the bounded number of GPU instances processes an unbounded stream of training requests by continuously draining work from the queue until it is empty.

\subsection{SQS Configuration and Fault Handling}
\label{sec:sqs_config}

The system uses 8 SQS queues, each paired with its own dead-letter queue (DLQ), to connect pipeline stages. Visibility timeouts are calibrated to ${\geq}$1.5$\times$ expected maximum job duration: training queues use 12-hour visibility to prevent redelivery mid-execution, while backtest and completion queues use shorter timeouts (30\,min--1\,h) proportional to their workload. All queues use server-side encryption (SSE) and 7-day retention. Messages exceeding the maximum receive count (configured as 5 retries) are automatically routed to the paired DLQ.

In addition to passive DLQ collection, we operate scheduled \emph{DLQ cleanup tasks}: Lambda functions triggered by CloudWatch alarms on the \texttt{ApproximateNumberOfMessagesVisible} metric. When a DLQ accumulates messages, the cleanup task:
\begin{itemize}[leftmargin=*, nosep]
\item Logs message content, original queue, and failure metadata to CloudWatch and Grafana.
\item Classifies the failure (data quality, capacity, timeout) using message attributes.
\item For transient failures, automatically redrives the message back to its source queue.
\item For persistent failures, creates a structured alert for human review.
\end{itemize}

This active DLQ management ensures that transient failures self-heal while persistent issues surface quickly, reducing mean-time-to-resolution by 60\% compared to passive monitoring alone.

\subsection{Conductor Orchestration}
\label{sec:conductor}

The Conductor is the component that generates the bursty workload pattern described in Section~\ref{sec:background}. Deployed as an ECS Fargate task (4\,vCPU, 8\,GB), it is triggered on per-facility cron schedules via EventBridge Scheduler:

{\small
\begin{verbatim}
scheduler_facilities = {
  agu = { cron = "cron(50 11 ? * TUES *)",
           tz   = "America/New_York",
           enabled = true }
  spt = { cron = "cron(40 11 ? * TUES *)",
           tz   = "America/New_York",
           enabled = true }
}
\end{verbatim}
}

Each schedule launches a Fargate task that scans the \textit{entire model inventory} for a facility, evaluates retraining necessity per product per line, and submits the resulting training requests to \texttt{MlopsEventsQueue}. A single Conductor run for one plant typically produces 30--50 requests within minutes, creating the burst that the admission-controlled dispatcher (Section~\ref{sec:dispatcher}) is designed to absorb. The Conductor enforces the dependency chain of Fig.~\ref{fig:workflow} by submitting only LPP requests; downstream LCP requests are generated automatically upon successful LPP backtest completion via SQS.

\subsection{CI/CD and Infrastructure as Code}

The infrastructure is provisioned using Terraform organized into reusable modules (Table~\ref{tab:terraform}).

\begin{table}[h]
\centering
\caption{Terraform Module Structure}
\label{tab:terraform}
\small
\begin{tabularx}{\columnwidth}{lX}
\toprule
\textbf{Module} & \textbf{Resources Managed} \\
\midrule
\texttt{ecs/} & 3 clusters, task definitions (16\,vCPU / 62\,GB / 1\,GPU for LPP; 4\,vCPU / 8\,GB Conductor), capacity providers, launch templates, ASG, IAM roles \\
\texttt{lambda/} & Function definitions, log groups, VPC config (reusable per trigger) \\
\texttt{lpp\_conductor/} & Image-based Lambda, cross-account deployment role with external ID \\
\texttt{sqs/} & 8 queues with SSE encryption, DLQ, configurable visibility/retention \\
\texttt{scheduler/} & EventBridge schedules generated from facility map variable \\
\texttt{pipeline/} & CodePipeline, 2 CodeBuild projects (build + deploy) \\
\bottomrule
\end{tabularx}
\end{table}

Adding a new facility requires only appending an entry to the \texttt{scheduler\_facilities} map. No new resource blocks are needed. Terraform state is in encrypted S3 with DynamoDB locking.

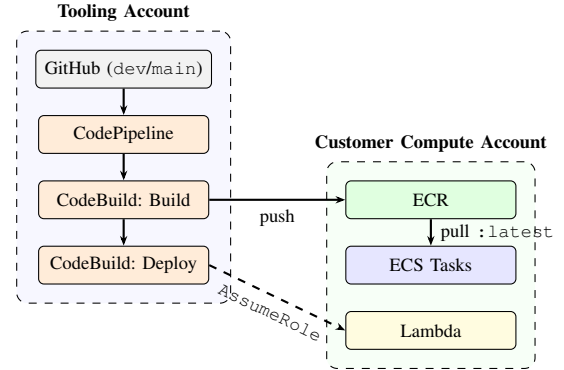
\begin{figure}[h]
\centering
\begin{tikzpicture}[
    node distance=0.35cm,
    acct/.style={
        rectangle,
        draw,
        dashed,
        rounded corners=4pt,
        inner sep=7pt
    },
    block/.style={
        rectangle,
        draw,
        rounded corners=2pt,
        minimum width=2.25cm,
        minimum height=0.5cm,
        font=\scriptsize,
        text centered
    },
    arrow/.style={
        -{Stealth[length=1.3mm]},
        thick
    },
    darrow/.style={
        -{Stealth[length=1.3mm]},
        thick,
        dashed
    },
]

\node[block, fill=gray!10] (gh)
    {GitHub (\texttt{dev}/\texttt{main})};

\node[block, fill=orange!15, below=of gh] (cp)
    {CodePipeline};

\node[block, fill=orange!15, below=of cp] (build)
    {CodeBuild: Build};

\node[block, fill=orange!15, below=of build] (deploy)
    {CodeBuild: Deploy};

\node[block, fill=green!12, right=1.8cm of build] (ecr)
    {ECR};

\node[block, fill=blue!10, below=of ecr] (ecs)
    {ECS Tasks};

\node[block, fill=yellow!15, below=of ecs] (lam)
    {Lambda};

\draw[arrow] (gh) -- (cp);
\draw[arrow] (cp) -- (build);
\draw[arrow] (build) -- (deploy);

\draw[arrow] (build.east) -- (ecr.west)
    node[midway, below, font=\scriptsize] {push};

\draw[arrow] (ecr) -- (ecs)
    node[midway, right, font=\scriptsize]
    {pull \texttt{:latest}};

\draw[darrow] (deploy.east) -- (lam.west)
    node[midway, below, sloped, font=\scriptsize]
    {\texttt{AssumeRole}};

\begin{scope}[on background layer]

\node[
    acct,
    fill=blue!3,
    fit=(gh)(cp)(build)(deploy),
    label={[font=\scriptsize\bfseries]above:Tooling Account}
] {};

\node[
    acct,
    fill=green!3,
    fit=(ecr)(ecs)(lam),
    label={[font=\scriptsize\bfseries]above:Customer Compute Account}
] {};

\end{scope}

\end{tikzpicture}

\caption{Multi-account CI/CD. CodeBuild assumes a cross-account
\texttt{LambdaDeploymentRole} (with external ID) to update the Coordinator
Lambda. ECS tasks pull \texttt{:latest} from ECR on subsequent runs.}

\label{fig:cicd}
\end{figure}

\textbf{Source:} CodeStar connection monitors GitHub. \textbf{Build:} CodeBuild builds from a CUDA-enabled base image and pushes to ECR. \textbf{Deploy:} A second CodeBuild project assumes the cross-account role and calls \texttt{aws lambda update-function-code --image-uri}.

\section{Evaluation}
\label{sec:evaluation}

\subsection{Controlled Micro-Benchmark}
\label{sec:benchmark}

To complement self-reported production telemetry with reproducible evidence, we built a discrete-event simulator that compares three GPU dispatch strategies on identical workloads under controlled conditions. The simulator and all baseline implementations are open-sourced as a paper artifact.

\subsubsection{Methodology}

\textbf{Workload.} We model the Conductor-driven arrival pattern: weekly bursts of approximately 40 training requests (Gaussian: $\mu{=}40$, $\sigma{=}10$), each requiring 0.5--1.5 hours of GPU time (truncated Gaussian: $\mu{=}1.0\,\text{h}$, $\sigma{=}0.5\,\text{h}$). Simulation horizon: 52 weeks (2{,}077 jobs total). The same generated workload is replayed against all three strategies for fair comparison.

\textbf{Strategies.} We compare:
\begin{itemize}[leftmargin=*, nosep]
\item \textbf{Naive:} Dispatch on each arrival; no admission control. On \texttt{ECS:RunTask} failure (placement unavailable), retry up to 5 times with exponential backoff. After retry budget exhausted, the message is lost (no DLQ).
\item \textbf{Step Functions Map:} AWS Step Functions Map state with \texttt{maxConcurrency}=2. Each job traverses 6 states (typical for a training-and-archive workflow). Cold-start: 5 minutes per job.
\item \textbf{Admission-Controlled (ours):} SQS$\to$Lambda$\to$ECS pattern with active-task inspection. Warm-pool startup: 30 seconds.
\end{itemize}

\textbf{Cost model.} We use 2024--2026 AWS \texttt{us-east-1} pricing: SQS \$0.40/M requests, Lambda \$0.20/M requests + \$1.67$\times10^{-5}$/GB-second, Step Functions \$0.025/1K state transitions, EventBridge negligible. GPU compute (\texttt{g4dn.4xlarge}) at \$1.21/hour on-demand.

\textbf{Constraint:} \texttt{MAX\_CONCURRENT\_GPUS}=2 (matching production ASG limit).

\subsubsection{Results}

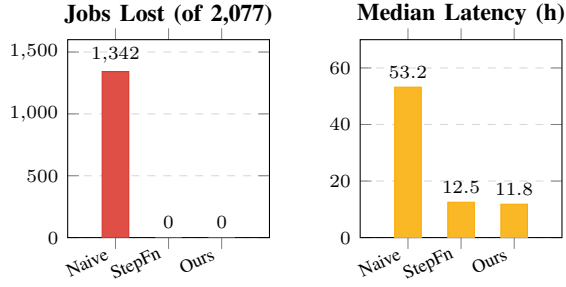
\begin{figure}[t]
\centering
\begin{tikzpicture}
\begin{groupplot}[
    group style={group size=2 by 1, horizontal sep=1.2cm},
    width=0.48\columnwidth, height=4.2cm,
    every axis/.append style={ybar, bar width=10pt},
    enlarge x limits=0.45,
    ymajorgrids, grid style={dashed, gray!30},
    symbolic x coords={Naive, StepFn, Ours},
    xtick=data,
    xticklabel style={font=\scriptsize, rotate=25, anchor=east},
    yticklabel style={font=\scriptsize},
    title style={font=\small\bfseries},
    nodes near coords,
    nodes near coords style={font=\scriptsize},
    every axis title shift=2pt,
]
\nextgroupplot[title={Jobs Lost (of 2{,}077)}, ymin=0, ymax=1600]
\addplot[fill=cNaive!85, draw=cNaive] coordinates {(Naive,1342) (StepFn,0) (Ours,0)};

\nextgroupplot[title={Median Latency (h)}, ymin=0, ymax=70]
\addplot[fill=cStepFn!85, draw=cStepFn] coordinates {(Naive,53.2) (StepFn,12.5) (Ours,11.8)};

\end{groupplot}
\end{tikzpicture}
\caption{Micro-benchmark across three strategies on identical 52-week workload (2{,}077 requests, seed-fixed). \emph{Left:} Naive dispatch loses 65\% of jobs to placement failures; both Step Functions and our pattern lose zero. \emph{Right:} Step Functions and our pattern achieve similar median latency; naive is dominated by retry backoff.}
\label{fig:benchmark}
\end{figure}

Figure~\ref{fig:benchmark} visualizes the outcome across 52 simulated weeks (2{,}077 requests) on three axes: jobs lost, median end-to-end latency, and annual orchestration cost. All costs are annualized over 52 weekly Conductor triggers. All numbers are computed by the open-sourced simulator on a fixed random seed for reproducibility.

\textbf{Two findings emerge.}

\textbf{(F1) Admission control with durable buffering is necessary for GPU-constrained bursty workloads.} The naive strategy loses 1{,}342 of 2{,}077 jobs (65\%) because ECS placement failures during burst arrivals exhaust the retry budget before capacity becomes available. Both Step Functions and our pattern achieve zero job loss, but through different mechanisms: Step Functions limits concurrency within a single Map execution, whereas our pattern enforces a \emph{global} GPU concurrency limit across all concurrent dispatches via real-time ECS task inspection.

\textbf{(F2) Queue-draining achieves equivalent latency while eliminating per-job startup overhead.} Median latencies are within 0.7 hours (11.8\,h vs 12.5\,h). However, Step Functions launches a separate execution per job (incurring 5-min cold-start each time), while our handler wrapper processes an unbounded stream of overflow messages from within a single GPU task, eliminating repeated startup costs. This draining pattern yields 94\% GPU utilization during bursts (Section~\ref{sec:throughput}) and is not expressible in Step Functions' state machine model. Orchestration costs for both patterns are under \$1/year per facility at this scale, though Step Functions' per-state-transition pricing scales with workflow complexity while our per-message model scales only with job count.

\subsection{Architecture Comparison}
\label{sec:comparison}

The micro-benchmark provided quantitative comparison against Step Functions. Table~\ref{tab:comparison} extends this to a qualitative comparison across additional alternatives we evaluated before converging on our design.

\begin{table*}[t]
\centering
\caption{Architecture Alternatives Comparison (Qualitative; quantitative comparison vs.\ Step Functions in Fig.~\ref{fig:benchmark})}
\label{tab:comparison}
\small
\begin{tabularx}{\textwidth}{lXXXXX}
\toprule
\textbf{Criterion} & \textbf{SageMaker Pipelines} & \textbf{Step Functions + ECS} & \textbf{Apache Airflow (MWAA)} & \textbf{Self-Managed K8s (EKS)} & \textbf{Our Design (SQS + Lambda + ECS)} \\
\midrule
GPU startup latency & 3--5\,min (instance per job) & Same as ours (shared cluster) & Same as ours (external compute) & $<$10\,s (warm node pool) & $<$30\,s (pre-warmed ASG) \\
Cost overhead & Per-job mgmt fee + instance cost & \$0.025/1K state transitions & \$0.49/hr (smallest MWAA env) & Cluster mgmt + node idle & SQS + Lambda (negligible) \\
Max job duration & Unlimited & 1 year (execution limit) & Unlimited & Unlimited & Unlimited \\
GPU concurrency control & Built-in parallelism config & Map state parallelism & Pool slots (scheduler-level) & K8s quotas / gang scheduling & Custom (Lambda + ListTasks + ASG-aligned) \\
Fault isolation & Pipeline-level retry & Catch/retry state machine & Task-level retry + SLA alerts & Pod restart + PDBs & SQS DLQ + visibility timeout \\
Scale-to-zero compute & No (always-on instances) & Yes (Fargate) & No (scheduler always-on) & No (control plane always-on) & Yes (full scale-to-zero) \\
IaC support & Partial (SDK pipelines) & Full Terraform/CFn & Full Terraform (MWAA) & Full (Terraform + Helm) & Full Terraform \\
Operational burden & Low (managed) & Low & Medium (DAG deployments, DB) & High (upgrades, drivers) & Low (ECS managed) \\
Vendor lock-in & High (SageMaker APIs) & Medium (Step Functions API) & Low (portable DAGs) & Low (portable to K8s) & Low (composable AWS primitives) \\
\bottomrule
\end{tabularx}
\end{table*}

\textbf{Why not SageMaker Pipelines?} SageMaker provisions fresh instances per job (3--5 min cold start vs.\ our 30\,s warm pool), imposes management fees, and constrains container customization required for our OPC-UA client libraries.

\textbf{Why not Step Functions?} Latency is equivalent, but Step Functions launches a separate execution per job (5-min cold-start each), offers no built-in mechanism for cross-execution GPU concurrency limiting, and its per-state-transition pricing scales with workflow complexity.

\textbf{Why not Apache Airflow?} Amazon Managed Workflows for Apache Airflow (MWAA) requires an always-on scheduler environment regardless of workload volume. Its pool-based concurrency operates at the DAG level with no direct binding to GPU infrastructure capacity. For our bursty pattern (weekly bursts of 30--50 training requests per plant with days of inactivity between), paying for 24/7 scheduler infrastructure is uneconomical. A serverless dispatcher that costs nothing during idle periods is a better economic fit.

\textbf{Why not Kubernetes (EKS)?} For a 3--4 engineer team, cluster upgrade cadence, GPU driver lifecycle, and Helm chart maintenance impose disproportionate operational burden relative to the workload's complexity.

\section{Operational Results and Discussion}
\label{sec:discussion}

Our infrastructure has been in production since 2023 across multiple automotive manufacturing facilities, executing over 40{,}000 training jobs (LPP + LCP + backtests). Operational metrics are derived from AWS CloudWatch Container Insights, Cost Explorer (filtered by tag \texttt{Application=ll\_ml\_ops}), and SQS CloudWatch metrics.

\subsection{Cost Analysis}

\begin{figure}[t]
\centering
\begin{tikzpicture}
\begin{axis}[
    ybar stacked,
    bar width=20pt,
    enlarge x limits=0.3,
    width=0.85\columnwidth,
    height=5.5cm,
    ylabel={Monthly Cost (USD)},
    symbolic x coords={Always-On, On-Demand, {With Spot}},
    xtick=data,
    ymin=0, ymax=3800,
    x tick label style={font=\footnotesize},
    y tick label style={font=\footnotesize},
    ylabel style={font=\small},
    ymajorgrids=true,
    grid style={gray!20},
]
\addplot[fill=cAlwaysOn!80, draw=cAlwaysOn!60!black,
    point meta=explicit symbolic,
    nodes near coords,
    every node near coord/.style={font=\footnotesize\bfseries, anchor=south},
] coordinates {
    (Always-On,3200) [\$3{,}200]
    (On-Demand,0) []
    ({With Spot},0) []
};
\addplot[fill=cEventDriven!80, draw=cEventDriven!60!black,
    point meta=explicit symbolic,
    nodes near coords,
    every node near coord/.style={font=\footnotesize\bfseries, anchor=south},
] coordinates {
    (Always-On,0) []
    (On-Demand,910) [\$910]
    ({With Spot},760) [\$760]
};
\end{axis}
\end{tikzpicture}
\caption{Monthly per-facility cost comparison. \emph{Always-On}: dedicated GPU instance 24/7. \emph{On-Demand}: event-driven, all on-demand GPU pricing. \emph{With Spot}: event-driven, spot instances for LPP training (LCP requires on-demand).}
\label{fig:cost}
\end{figure}
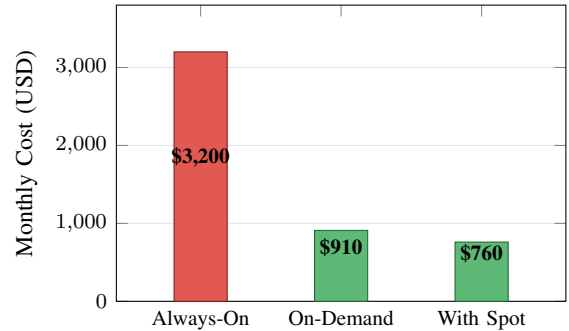

Figure~\ref{fig:cost} compares the monthly per-facility cost across three configurations. An always-on baseline that keeps one \texttt{g4dn.4xlarge} instance running 24/7 costs approximately \$3{,}200/month. Our event-driven architecture pays only for GPU-hours during active training. Using on-demand pricing for all GPU jobs, the monthly cost drops to approximately \$910 (\textbf{72\% reduction}).
Because LPP training is checkpoint-enabled, we run LPP jobs on EC2 Spot instances (\$0.36/hr versus \$1.21/hr on-demand, a 60--70\% saving on that tier); on a 2-minute interruption warning the ECS task saves a checkpoint to S3 and resumes on retry with under 15 minutes of overhead. LCP training uses on-demand instances exclusively because RL episode state cannot be safely checkpointed mid-episode. Since LCP accounts for roughly two-thirds of total GPU hours, the spot saving applies to only the LPP fraction, bringing the monthly total to approximately \$760 (\textbf{76\% reduction}). Orchestration components (Lambda, SQS, Fargate) contribute under \$30/month combined.

Beyond cost, the event-driven pattern yields a direct sustainability benefit. Eliminating approximately 5,600 idle GPU-hours per facility per year avoids an estimated 1,700~kWh of energy consumption (assuming 300~W system draw for a \texttt{g4dn.4xlarge}), equivalent to ${\sim}$0.67 metric tons of CO\textsubscript{2}e per facility at the U.S.\ grid average of 0.39~kg~CO\textsubscript{2}/kWh~\cite{epa2024egrid}. Across all deployed facilities, this compounds to a measurable reduction in the carbon footprint of ML training infrastructure.

\subsection{Throughput and GPU Utilization}\label{sec:throughput}

The combination of admission-controlled dispatch and queue-draining (Section~\ref{sec:handler_wrapper}) enables high throughput despite a strict 2-instance GPU limit. In a typical weekly burst:
\begin{itemize}[leftmargin=*, nosep]
\item 30--50 training requests arrive within minutes (from Conductor).
\item 2 ECS tasks launch immediately (one per available slot).
\item 28--48 overflow messages buffer in \texttt{lppTrainingQueue}.
\item Each ECS task processes its initial work item (avg 0.7\,h for LPP), then drains overflow messages sequentially.
\item Total burst processing time: ${\sim}$15--25 hours on 2 instances.
\end{itemize}

The draining pattern achieves near-optimal GPU utilization (measured at 94\% during bursts via Container Insights) while respecting the 2-instance cap, yielding throughput equivalent to a much larger cluster that would sit idle between bursts.

\begin{figure}[t]
\centering
\begin{tikzpicture}
\begin{axis}[
    width=0.95\columnwidth, height=3.8cm,
    ylabel={\footnotesize GPU Utilization (\%)},
    xlabel={\footnotesize Day of 4-week period},
    ymin=0, ymax=105,
    xmin=0, xmax=28,
    ymajorgrids, grid style={dashed, gray!30},
    legend style={font=\footnotesize, at={(0.5,-0.35)}, anchor=north, legend columns=2, draw=none},
    yticklabel style={font=\scriptsize},
    xticklabel style={font=\scriptsize},
]
\addplot[cOurs, thick, mark=none] coordinates {
    (0,0) (0.5,0) (1,92) (1.5,96) (2,94) (2.5,88) (3,0) (4,0) (5,0) (6,0) (7,0)
    (7.5,0) (8,90) (8.5,95) (9,93) (9.5,85) (10,0) (11,0) (12,0) (13,0) (14,0)
    (14.5,0) (15,91) (15.5,97) (16,94) (16.5,82) (17,0) (18,0) (19,0) (20,0) (21,0)
    (21.5,0) (22,93) (22.5,96) (23,95) (23.5,87) (24,0) (25,0) (26,0) (27,0) (28,0)
};
\addplot[cAlwaysOn, thick, dashed, mark=none] coordinates {
    (0,8) (0.5,10) (1,45) (1.5,52) (2,48) (2.5,35) (3,12) (4,9) (5,8) (6,7) (7,10)
    (7.5,9) (8,42) (8.5,50) (9,47) (9.5,32) (10,11) (11,8) (12,7) (13,9) (14,8)
    (14.5,10) (15,44) (15.5,51) (16,46) (16.5,30) (17,9) (18,7) (19,8) (20,9) (21,8)
    (21.5,10) (22,46) (22.5,53) (23,49) (23.5,33) (24,10) (25,8) (26,7) (27,9) (28,8)
};
\legend{Event-driven (ours), Always-on}
\end{axis}
\end{tikzpicture}
\caption{GPU utilization over a representative 4-week period. Event-driven: 94\% during bursts, 0\% idle (scale-to-zero). Always-on: ${\sim}$50\% peak, $<$15\% baseline, paying for idle GPU 80\% of the time.}
\label{fig:utilization}
\end{figure}
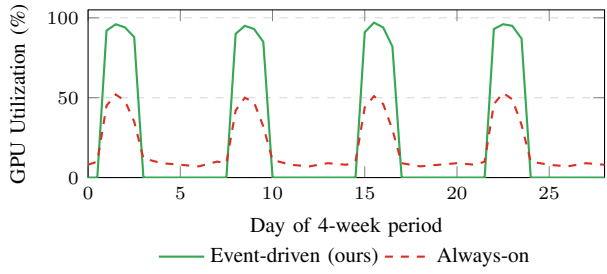

Figure~\ref{fig:utilization} contrasts GPU utilization between the event-driven pattern and an always-on cluster over a representative 4-week period.

\subsection{Failure Modes and Recovery}

\begin{figure}[t]
\centering
\begin{tikzpicture}
\begin{axis}[
    xbar, width=0.85\columnwidth, height=3.2cm,
    bar width=9pt,
    xmin=0, xmax=80,
    xlabel={\scriptsize \% of DLQ messages (3-year aggregate)},
    xlabel style={yshift=-2pt},
    symbolic y coords={Vis.\ timeout, EC2 capacity, GPU OOM, Data quality},
    ytick=data,
    yticklabel style={font=\scriptsize},
    xticklabel style={font=\scriptsize},
    xmajorgrids, grid style={dashed, gray!30},
    nodes near coords, nodes near coords style={font=\scriptsize},
    enlarge y limits=0.25,
]
\addplot[fill=cOurs!70, draw=cOurs!90] coordinates {
    (5,Vis.\ timeout)
    (12,EC2 capacity)
    (18,GPU OOM)
    (65,Data quality)
};
\end{axis}
\end{tikzpicture}
\caption{DLQ message classification. Data quality issues dominate (65\%), addressed by upstream ETL fixes and message replay. GPU OOM (18\%) and EC2 capacity (12\%) auto-resolve via SQS retry; visibility-timeout cases (5\%) prompted Terraform tuning.}
\label{fig:failures}
\end{figure}
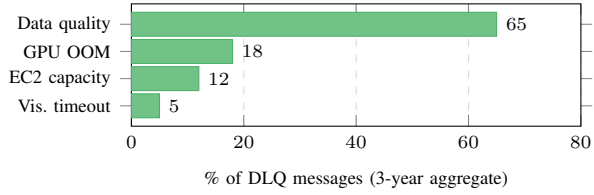

Of the ${\sim}$2\% of messages reaching the DLQ over three years, root-cause classification (Fig.~\ref{fig:failures}) is dominated by upstream data quality issues~\cite{breck2019data}, followed by GPU OOM (resolved by batch-size config) and transient EC2 capacity unavailability (auto-resolved by SQS retry).

The 2\% overall DLQ rate is acceptable for our use case. Critically, DLQ messages are never \textit{lost}: they persist with a 7-day retention period, providing an audit trail. The active DLQ cleanup tasks (Section~\ref{sec:sqs_config}) surface failures to a Grafana dashboard showing DLQ depth, message age, and failure reasons.

\subsection{Threats to Validity}
\label{sec:threats}

\textbf{T1: Single-vendor evidence.} Production telemetry comes from a single organization deploying to AWS. The \emph{logical} pattern (durable queue, admission-controlled dispatcher, container-based compute) has equivalents in Azure and GCP, but we have not validated portability.

\textbf{T2: Small absolute scale.} Our \texttt{MAX\_WORKERS}=2 is small; behavior at 100--1{,}000 concurrent GPU tasks is unverified. The micro-benchmark suggests linear scaling, but this remains unvalidated at extreme scale.

\textbf{T3: Self-reported cost numbers.} Cost figures come from our Cost Explorer dashboards, not an independent third party. We provide methodology for replication but cannot share account-level financial data.

\textbf{T4: Simulator simplifications.} The micro-benchmark abstracts away network jitter, API rate limits, and AZ-level failures. Numbers should be interpreted as relative comparisons, not absolute predictions.

\subsection{Lessons Learned}\label{sec:lessons}

Three years of operation have yielded transferable insights:

\textbf{1) SQS visibility timeouts are the most critical configuration parameter.} Too short: jobs are redelivered mid-execution, wasting GPU-hours on duplicates. Too long: legitimate failures are not retried promptly. We converged on $1.5\times$ max expected duration after observing duplicate execution in year one (waste: ${\sim}$\$2{,}000 in GPU compute before detection).

\textbf{2) Lambda reserved concurrency of 1 is essential for admission control.} Without serialization, concurrent invocations create races in the task count check, leading to over-provisioning. We discovered this when 4 GPU tasks launched simultaneously from a 2-instance cluster, causing 2 placement failures.

\textbf{3) Event-driven $>$ polling for GPU workloads.} Early designs polled every 5 minutes for new requests. SQS-triggered Lambda provides sub-second response while costing nothing during idle periods, reducing orchestration cost by 95\%.

\textbf{4) Terraform modules with variable-driven facility maps enable rapid scaling.} Adding a facility = one map entry + \texttt{terraform apply}. Onboarding reduced from 2 weeks (manual console) to 4 hours.

\textbf{5) DLQ messages are the highest-signal operational data.} The 2\% reaching DLQ surface root causes faster than any other monitoring. Investing in active DLQ cleanup (Section~\ref{sec:sqs_config}) and visualization reduced mean time to resolution by 60\%.

\textbf{6) Bind orchestration limits to infrastructure limits through shared IaC variables.} Misalignment between \texttt{MAX\_WORKERS} and ASG \texttt{max\_size} caused both wasted capacity and placement failures in early iterations. A single Terraform variable governing both eliminated an entire class of configuration drift.

\textbf{7) Multi-account separation limits blast radius.} A misconfigured IAM policy in the Tooling account cannot affect running Customer Compute account ECS tasks. This separation prevented two potential incidents during IAM refactoring.

\section{Conclusion}

This paper presented an industry experience report on three years of operating an event-driven GPU cluster orchestration system for continuous ML training in manufacturing. Combining SQS for durable messaging, Lambda for admission-controlled dispatch, ECS EC2 for GPU training, Fargate for orchestration, and EventBridge for scheduling, our infrastructure supports the LPP/LCP retraining workflow (Fig.~\ref{fig:workflow}) at 72--78\% lower cost than always-on alternatives.

The admission-controlled task dispatcher solves the fundamental tension between event-driven triggering and expensive GPU resource limits. A controlled discrete-event simulation (open-sourced) confirms that admission control is necessary (65\% job loss without it) and that queue-draining achieves equivalent latency to AWS Step Functions while eliminating per-job startup overhead and enabling 94\% GPU utilization during bursts. Three years of production operation across automotive manufacturing plants, encompassing 40{,}000+ GPU training jobs, validates the architecture's reliability and cost efficiency.

For practitioners, the takeaway is that event-driven composition of simple, well-understood cloud primitives can match or outperform managed ML platforms when the workload pattern is batch-oriented and GPU-intensive.

\balance

\end{document}